\documentclass[journal]{IEEEtran}
\usepackage{ulem}
\ifCLASSINFOpdf
\else
\fi

\usepackage{caption}
\usepackage{makecell}
\usepackage{graphicx,wrapfig,fullpage,amsmath,hhline,epsfig,verbatim,url,amssymb,multicol,multirow,cite}
\usepackage{times,color,soul} 
\usepackage[left=0.75in,top=0.70in,right=0.75in,bottom=0.80in]{geometry}
\usepackage{amsbsy,latexsym,mathrsfs,mathtools}

\usepackage{mathptmx} 
\DeclareMathAlphabet{\mathcal}{OMS}{cmsy}{m}{n}
\usepackage{bm}
\usepackage{float}
\usepackage{color,soul}
\usepackage{dsfont}
\usepackage{comment}
\usepackage{algorithm}
\usepackage{algorithmic}
\usepackage{psfrag} 
\usepackage[table,dvipsnames]{xcolor}

\usepackage{url}
\definecolor{darkgreen}{rgb}{0,0.7,0.1}
\definecolor{darkblue}{rgb}{0.1,0.3,0.8}
\definecolor{darkred}{rgb}{0.6,0,0}
\usepackage[bookmarks=true,colorlinks=true,pdfpagemode=UseNone,citecolor=darkgreen,linkcolor=darkblue,urlcolor=blue]{hyperref}
\usepackage{setspace}
\usepackage{paralist}
\usepackage{booktabs}
\usepackage{balance}

\usepackage[switch]{lineno}

\definecolor{lightblue}{rgb}{0.9, 0.95, 1}

\title{\LARGE \bf
    Model-agnostic Meta-learning for Adaptive Gait Phase and Terrain Geometry Estimation with Wearable Soft Sensors
}

\author{Zenan Zhu$^{1,*}$, 
    Wenxi Chen$^{1,*}$,
    Pei-Chun~Kao$^{2}$,
    Janelle Clark$^{4}$,
    Lily~Behnke$^{3}$, \\
    Rebecca~Kramer-Bottiglio$^{3}$,
    ~Holly~Yanco$^{2}$, 
    Yan Gu$^{1,\dagger}$

\thanks{$^{1}$Z. Zhu, W. Chen, and Y. Gu are with the School of Mechanical Engineering, Purdue University, West Lafayette, IN 47907, USA (e-mail: \{zhu1134, chen1318, yangu\}@purdue.edu).
$^{2}$P. Kao is with the Dept. of Physical Therapy and Kinesiology, and H. Yanco is with the Miner School of Computer \& Information Sciences, University of Massachusetts Lowell, Lowell, MA 01854, USA.
$^{3}$L. Behnke and R. Kramer-Bottiglio are with the Dept. of Mechanical Engineering \& Materials Science, Yale University, New Haven, CT 06520, USA.
$^{4}$J. Clark is with the Dept. of Mechanical Engineering, University of Maryland Baltimore County, Baltimore, MD 21250, USA.
This research has been supported in part by NSF IIS-1955979.
$^{*}$These authors contributed equally.
$^{\dagger}$Corresponding author.}
}

\begin{document}	
\maketitle
\thispagestyle{plain}
\pagestyle{plain}

\begin{abstract}

    This letter presents a model-agnostic meta-learning (MAML) based framework for simultaneous and accurate estimation of human gait phase and terrain geometry using a small set of fabric-based wearable soft sensors, with efficient adaptation to unseen subjects and strong generalization across different subjects and terrains. 
    Compared to rigid alternatives such as inertial measurement units, fabric-based soft sensors improve comfort but introduce nonlinearities due to hysteresis, placement error, and fabric deformation. Moreover, inter-subject and inter-terrain variability, coupled with limited calibration data in real-world deployments, further complicate accurate estimation. 
    To address these challenges, the proposed framework integrates MAML into a deep learning architecture to learn a generalizable model initialization that captures subject- and terrain-invariant structure.
    This initialization enables efficient adaptation (i.e., adaptation with only a small amount of calibration data and a few fine-tuning steps) to new users, while maintaining strong generalization (i.e., high estimation accuracy across subjects and terrains).
    Experiments on nine participants walking at various speeds over five terrain conditions demonstrates that the proposed framework outperforms baseline approaches in estimating gait phase, locomotion mode, and incline angle, with superior accuracy, adaptation efficiency, and generalization.

\end{abstract}

\vspace{-0.15 in}
\section{Introduction}
\vspace{-0.05 in}
\label{sec: intro}

Accurate estimation of human movement and terrain geometry can support lower-limb wearable robots (e.g., exoskeletons) to provide adaptive assistance across diverse walking conditions~\cite{young2016state}. Gait phase identification enables synchronization of robotic assistance with biomechanical events such as ankle push-off~\cite{cortino2023data, chinimilli2019automatic}, while terrain-aware control modulates assistance based on terrain features (e.g., incline angle) to facilitate smooth transitions across terrains~\cite{cheng2025ambilateral}.

In real-world deployment, estimators must adapt rapidly to previously unseen users using only a small amount of subject-specific calibration data, while maintaining high estimation accuracy across different users and terrain conditions.
These requirements pose substantial challenges for conventional learning methods, which typically rely on large training datasets and suffer performance degradation under subject- or terrain-specific variability~\cite{alzubaidi2021review}.
To address these challenges, this study proposes a gait phase and terrain geometry estimation framework based on MAML, designed to enable efficient adaptation to new users while ensuring generalization across users and terrains.

\vspace{-0.15 in}
\subsection{Related Work}

\vspace{-0.05 in}

\subsubsection{Model-based estimation}
Model-based methods commonly employ kinematic models and sensor fusion techniques (e.g., dead-reckoning with foot-mounted inertial measurement units (IMUs)~\cite{hou2020pedestrian} and Kalman filters with biomechanical models~\cite{yuan20133}) for real-time human movement estimation. Yet, they often suffer from integration error accumulation, sensitivity to modeling assumptions, and uncertain limb geometry and sensor placement~\cite{zhang2015whole}.

\subsubsection{Learning-based estimation}

Machine learning methods (e.g., deep neural networks) have been widely adopted for human movement estimation and to address nonlinearities of soft wearable sensors~\cite{kim2018deep, li2023shortcut,molinaro2024task}.
Although humans can often adapt to slightly mismatched settings~\cite{xu2025reciprocal}, wearable systems that are fully tailored to a new user still require extensive recalibration data and fine-tuning. This is due to substantial variations in individual gait patterns, influenced by factors such as age, muscle strength, and neuromuscular control~\cite{kowalsky2021human,xu2024chatemg}.
Transfer learning can reduce data requirements by pre-training on able-bodied subjects and selectively fine-tuning for unseen target users (e.g., amputees)~\cite{le2024transfer};
still, the adaptation efficiency and generalization across unseen subjects may not be ensured due to distribution shift between training and test data.

\subsubsection{Meta-learning}

Compared to conventional learning-based estimation methods, meta-learning enables models to rapidly adapt to unseen conditions using prior experience, with significantly fewer calibration data and fine-tuning steps~\cite{hospedales2021meta}.
Recent studies have demonstrated meta-learning’s effectiveness for wearable sensing, including gait phase estimation from IMU data using stacked denoising autoencoders~\cite{cao2024fusion} and electromyography-based hand orthosis control~\cite{la2024meta}.
Yet, these methods are typically evaluated in limited scenarios—e.g., a few movement modes and a single environment—leaving open questions about their ability to generalize to more diverse subjects, movement patterns, and environmental contexts.

\subsubsection{Rigid vs. soft wearable sensors}

Prior methods of human movement and terrain geometry estimation typically use rigid sensors such as IMUs and motion-capture systems~\cite{zhu2022design, leestma2023linking}, which may be uncomfortable or impractical for continuous use.
In contrast, soft sensors are lightweight, compliant, and garment-integrable, making them suitable for daily monitoring~\cite{Walshwearablesensor}.
While most soft sensors in the literature and on the market are elastomer-based, recent fabric-based stretchable strain sensors offer enhanced comfort with improved air permeability and water vapor transmission~\cite{sanchez2023stretchable}.
Yet, these sensors exhibit inherent nonlinearities due to material properties, skin-garment shifts, and deformation from prolonged wear~\cite{case2019robotic, zhang2023intelligent}, complicating accurate estimation.

\vspace{-0.15 in}
\subsection{Contributions}
\vspace{-0.05 in}

This study introduces a MAML-based estimation framework using a small set of wearable soft sensors to accurately and simultaneously estimate human gait phase and terrain geometry.
The main contributions of this work are as follows:
\textbf{(a)} Construction of a deep convolutional neural network (DCNN) with multi-head structure to simultaneously estimate gait phase, locomotion mode, and incline angle from a small amount of fabric-based sensor data.
\textbf{(b)} Integration of MAML into the DCNN to enable rapid, data-efficient adaptation to unseen subjects and robust generalization across users and terrains.
\textbf{(c)} Extensive evaluation on nine participants walking at multiple speeds and terrain types, demonstrating that the proposed approach outperforms standard supervised and transfer learning baselines in accuracy, adaptation efficiency, and generalization.

\vspace{-0.10 in}
\section{Problem Formulation} \label{sec: problem formulation}
\vspace{-0.05 in}

This section formulates the problem of estimating gait phase and terrain geometry from wearable soft sensor data.

\vspace{-0.15 in}
\subsection{Gait Phase and Terrain Geometry Definition} 
\vspace{-0.05 in}
\label{sec: gait phase and terrain}
The problem formulation begins by defining the gait phase and terrain geometry variables to be estimated.

\subsubsection{Gait phase}
A complete human walking cycle consists of four phases, each defined by distinct foot-ground contact conditions~\cite{kharb2011review}, as shown in
Fig.\ref{fig: gait phase demo}.
These phases, denoted as G1-G4, are:
\textbf{G1} is a double-support phase with both feet on the ground and the right foot about to lift off; 
\textbf{G2} is a left-foot-single-support phase, with the left foot in contact and the right foot swinging;
\textbf{G3} is another double-support phase with the left foot about to lift off;
and
\textbf{G4} is a right-foot-single-support phase, with only the right foot in contact.

\subsubsection{Terrain geometry}
To reflect common real-world terrains, we classify five discrete locomotion modes: \textbf{level walking (LW)}, \textbf{ramp ascent (RA)}, \textbf{ramp descent (RD)}, \textbf{stair ascent (SA)}, and \textbf{stair descent (SD)}.
In addition to mode classification, the terrain slope angle is estimated to capture differences in incline across terrains.

\subsubsection{Estimation variables}

This study focuses on estimating three variables at each time step $t \in \mathbb{N_+}$: locomotion mode $l_t \in \{\text{LW, RA, RD, SA, SD}\}$, gait phase $g_t \in \{\text{G1, G2, G3, G4}\}$, and terrain incline angle $\phi_t \in \mathbb{R}$. 
Here, $(\cdot)_t$ denotes the value of the variable $(\cdot)$ at time $t$. 
Note that the locomotion mode \( l_t \) and gait phase \( g_t \) are discrete variables with symbolic labels, while the incline angle \( \phi_t \) is treated as a continuous variable.

\begin{figure}[t]
    \centering
    \includegraphics[width= 0.48\textwidth]{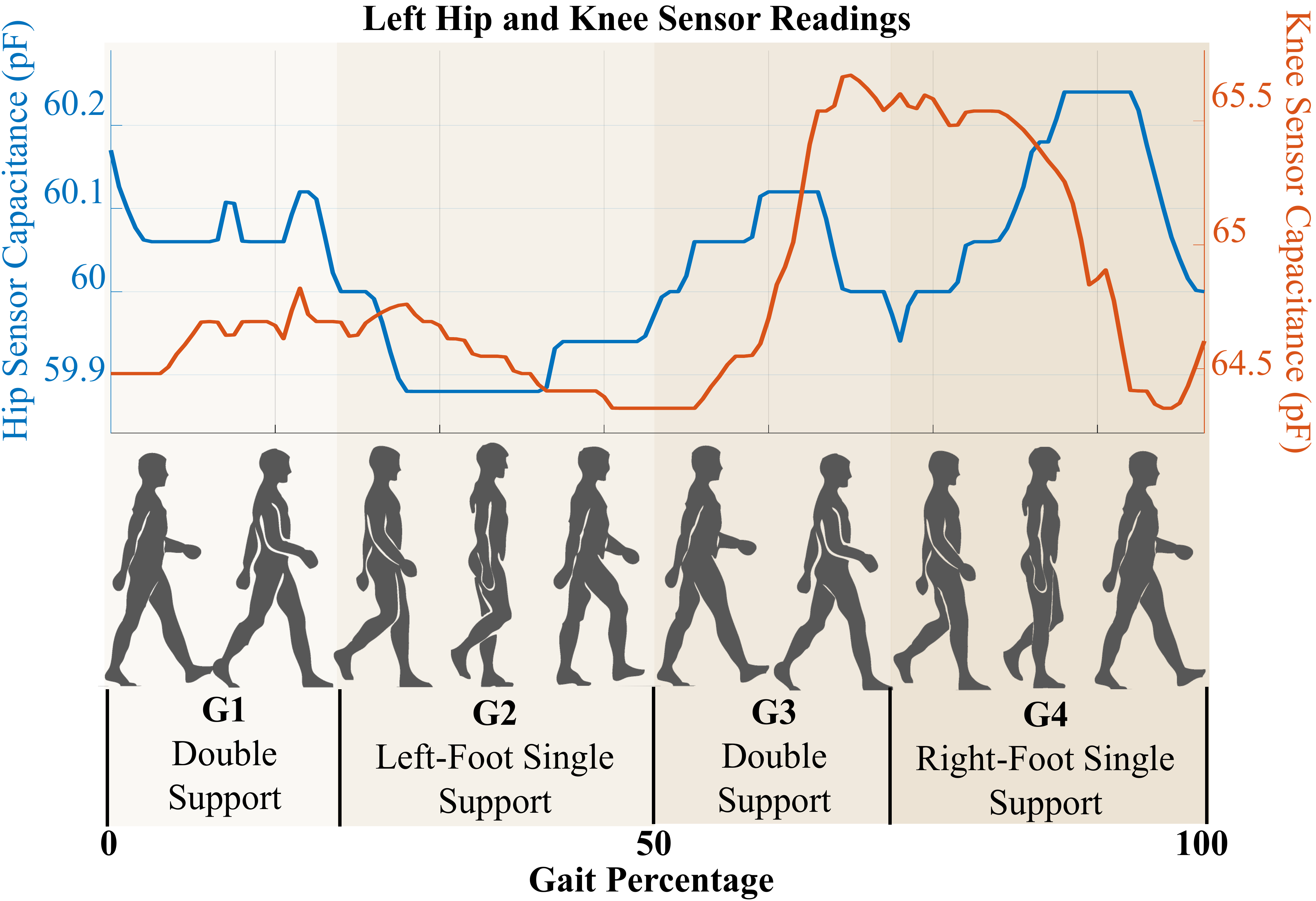}
    \vspace{-0.25 in}
    \caption{Example of walking gait phases along with representative soft sensor measurements.}
    \label{fig: gait phase demo}
\end{figure}

\vspace{-0.15 in}
\subsection{Sensors Considered}
\vspace{-0.05 in}
\label{sec: sensor considered}
To enhance wearer comfort compared to rigid sensors, this study uses data from a small set of fabric sensors~\cite{sanchez2023stretchable}, embedded in a pair of compression pants.
Four fabric sensors are placed along the sagittal planes at the hip and knee joints.
Each sensor comprises two dielectric layers (with one formed by the pants) and three electrode layers, all made of nylon.
Due to humidity and sweat sensitivity, the sensors were encapsulated in elastomer.
The encapsulated sensors are embedded in the garment, with the garment itself acting as one of the dielectric layers in the capacitor, such that, during quiet standing, they remain stretched.
As a hip or knee rotates, the corresponding sensor stretches, producing capacitance changes that reflect joint angle variations (Fig.\ref{fig: gait phase demo}).

\vspace{-0.15 in}
\subsection{Objective} 
\vspace{-0.05 in}

This study aims to achieve accurate and simultaneous estimation of locomotion mode $l_t$, gait phase $g_t$, and terrain incline angle $\phi_t$ using only a small set of wearable soft sensors, with efficient adaptation (i.e., rapid fine-tuning to unseen users using limited calibration data) and high generalization (i.e., consistently accurate estimation across locomotion modes and terrain geometry after efficient adaptation to unseen subjects).
A key challenge lies in the substantial variability across individuals and terrains, due to differences in sensor placement, movement patterns, terrain characteristics, and the inherent nonlinearities of soft sensors.

\vspace{-0.15 in}
\section{Methodology} \label{sec: methodology}
\vspace{-0.05 in}

This section introduces the proposed MAML-based framework for estimating gait phase and terrain geometry.

\begin{figure}[t]
    \centering
    \includegraphics[width= 0.48\textwidth]{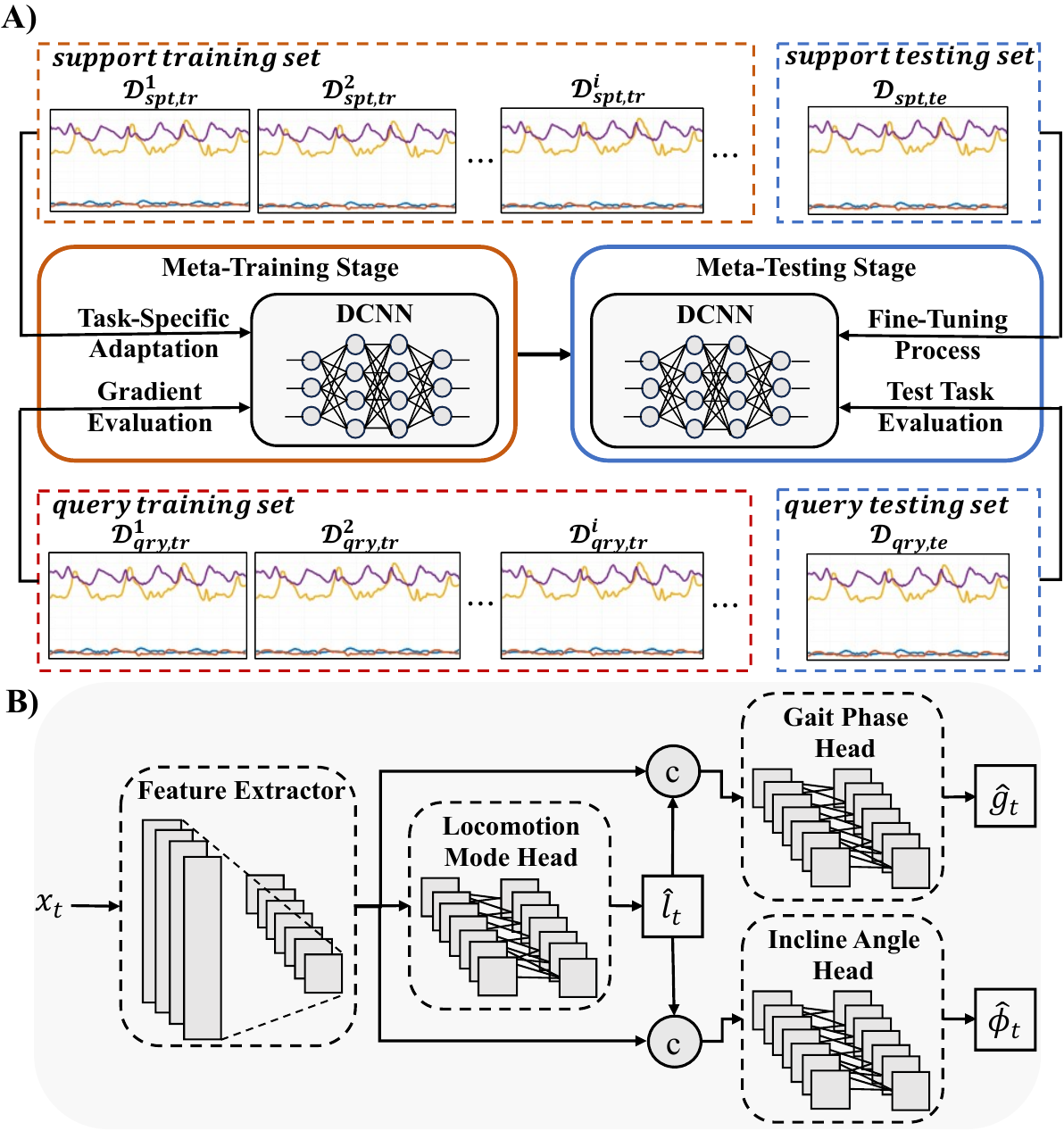}
    \vspace{-0.2 in}
    \caption{
    (A) MAML pipeline with task-specific adaptation and gradient aggregation during meta-training. (B) Shared DCNN architecture for meta-training and meta-testing.}
    \label{fig: maml pipeline}
\end{figure}

We choose to construct a DCNN (Fig.~\ref{fig: maml pipeline}(B)) as the backbone of the proposed framework, considering its ability to capture local temporal patterns in time-series data and support multi-output estimation
while also implicitly learning to compensate for the nonlinearity of soft sensor signals.

The MAML pipeline comprises two stages (Fig.~\ref{fig: maml pipeline}(A)): meta-training and meta-testing.
During meta-training, the DCNN model learns a generalizable initialization by performing task-specific adaptation and aggregating gradients across different subjects and terrains, enabling efficient adaptation to new conditions.
Meta-testing involves deploying this meta-trained model on unseen users, using limited calibration data for rapid fine-tuning.

Details of the proposed framework are explained next.

\vspace{-0.15 in}
\subsection{DCNN Model} \label{sec: inner model architecture}
\vspace{-0.05 in}

As introduced earlier, the DCNN model serves as the backbone of the proposed MAML-based estimation framework. The model adopts a multi-head architecture in which three estimation heads (i.e., locomotion mode, gait phase, and incline angle heads) share a common feature encoder, enabling efficient parameter sharing and reduced computational overhead.
A distinguishing feature of this design is that the predicted locomotion mode is concatenated with the shared features and used as input to the gait phase and incline angle heads. This reflects the biomechanical dependency of these variables and contributes towards accurate estimation while preserving computational efficiency.

\subsubsection{Model input}

To enable estimation with only a few soft sensors, we construct the model input at each time step $t \in \mathbb{N}_+$ from a sliding window of the 
previous $k$ frames from four sensor channels at the hips and knees (Sec.~\ref{sec: problem formulation}.B).
Formally, the input $x_t \in \mathbb{R}^{4 \times k}$ captures short-term motion characteristics.
In our experiments (Sec.~\ref{sec: experiment}), we use $k = 100$ (approx. one second), which typically spans a full gait cycle.

\subsubsection{Model output}
Since the proposed framework aims to estimate locomotion mode $l_t $, gait phase $g_t$, and terrain incline angle $\phi_t$ at each time step $t$ (Sec.~\ref{sec: problem formulation}.A.3), we define the DCNN model's output as $y_t:= \{l_t, g_t, \phi_t\}$.
Given input $x_t$, the DCNN model, parameterized by $\theta$ and denoted as $f_{\theta}$, predicts an estimate $\hat{y}_t$ of the output, i.e., $\hat{y}_t= f_{\theta}(x_{t}).$

\subsubsection{Feature extractor}
To ensure efficiency and prevent overfitting, a feature extractor is constructed with a one-dimensional convolutional layer for processing the temporal input signal ${x}_t$, followed by a batch normalization layer and a ReLU-activated linear layer.

\subsubsection{Multi-head structure}
To jointly estimate gait phase and terrain geometry (i.e., ${y}_t$), the DCNN employs a multi-head architecture (Fig.~\ref{fig: maml pipeline}(B)), comprising: (a) a locomotion mode classification head for $l_t$; (b) a gait phase classification head for $g_t$; and (c) an incline regression angle head for $\phi_t$.
Each head is a two-layer perceptron with ReLU activation.
To reduce computational cost, all heads share the same extracted features.
Also, as mentioned earlier, since locomotion mode affects joint kinematics, its prediction is concatenated with the shared features and fed into the other two heads.

\begin{algorithm}[t]
\caption{Meta-training of the proposed MAML-based state estimation approach}
\label{algorithm:meta-train}
\begin{algorithmic}[1]
\REQUIRE Data partitioned by traning tasks $\mathcal{T}_{\text{tr}}^i $ \\
\REQUIRE $\alpha$, $\beta$: step size hyperparameters
\STATE Randomly initialize $\theta$
\WHILE{not converged} 
    \FOR{all $\mathcal{T}_{\text{tr}}^i$}

        \STATE Sample training set $\mathcal{D}_{\text{tr}}^{i} \sim \mathcal{T}_{\text{tr}}^i$ and partition into disjoint support set $\mathcal{D}_{\text{spt},\text{tr}}^{i}$ and query set $\mathcal{D}_{\text{qry},\text{tr}}^{i}$

        \STATE Evaluate gradient of loss on support set \\ $\nabla_{\theta} \mathcal{L}(f_\theta;\mathcal{D}_{\text{spt},\text{tr}}^{i})$ 
        \STATE Compute adapted parameters with gradient descent: $\theta'_i = \theta - \alpha \nabla_{\theta} \mathcal{L}(f_{\theta};\mathcal{D}_{\text{spt},\text{tr}}^{i})$
        \STATE Evaluate loss on query set with adapted parameters $\mathcal{L}(f_{\theta'_i};\mathcal{D}_{\text{qry},\text{tr}}^{i})$
    \ENDFOR
    \STATE Update $\theta \leftarrow \theta - \beta \nabla_{\theta} \sum_{i} \mathcal{L}(f_{\theta'_i};\mathcal{D}_{\text{qry},\text{tr}}^{i})$
\ENDWHILE
\end{algorithmic}
\end{algorithm}

\vspace{-0.15 in}
\subsection{Meta-training Stage} \label{sec: meta train loop}
\vspace{-0.05 in}

The meta-training stage learns a DCNN model initialization that enables efficient adaptation to new conditions.

\subsubsection{Algorithm overview}

As summarized in Algorithm~\ref{algorithm:meta-train}, we partition the training data by different tasks.
Each training task $\mathcal{T}^i_{\text{tr}}$ corresponds to a walking session by a subject under a specific terrain condition and speed.
This task construction organizes the training data to reflect variations across subjects, terrains, and walking speeds, encouraging the model to learn task-invariant representations that generalize well across deployment scenarios.

Within each meta-training epoch (i.e., a single iteration of the \texttt{while} loop), two main steps are performed for each task:
(I) \textbf{task-specific adaptation}, where the model is fine-tuned on each task’s support set to obtain task-adapted parameters $\theta'_i$, and (II) \textbf{meta-update}, where the shared initialization $\theta$ is updated based on query set losses aggregated across tasks.

\subsubsection{Loss function design}
Meta-training uses a loss function $\mathcal{L}$, which we define as 
the weighted sum of the estimation errors of the gait phase, incline angle, and locomotion mode. The loss of model $f_{ \theta}$ evaluated on a data set $\mathcal{D}$ is:
\begin{equation}
\label{eq: weighted loss}
\begin{aligned}
\mathcal{L}\left(f_{ \theta}
; \mathcal{D}\right)
= &w_{\text{gait}}\,\mathcal{L}_{\text{gait}}\left(f_{ \theta}; \mathcal{D}\right)
+ w_{\text{inc}}\,\mathcal{L}_{\text{inc}}\left(f_{ \theta}; \mathcal{D}\right)\\
&+ w_{\text{loc}}\,\mathcal{L}_{\text{loc}}\left(f_{ \theta}; \mathcal{D}\right),
\end{aligned}
\end{equation}
where $w_{\text{gait}}$, $w_{\text{inc}}$, and $w_{\text{loc}}$ are predefined weighting factors, and $\mathcal{L}_{\text{gait}}$, $\mathcal{L}_{\text{inc}}$, and $\mathcal{L}_{\text{loc}}$ denote individual task losses.

For the classification of locomotion mode and gait phase and the regression associated with incline angle estimation, we use the commonly used cross-entropy loss and root-mean-square error (RMSE), respectively.
The same loss functions and weighting parameters are applied
for both meta-training and meta-testing, including fine-tuning steps.

\subsubsection{Sampling support and query sets (Line 4 of Algorithm~\ref{algorithm:meta-train})}
At each \texttt{for} loop iteration,
a training dataset $ \mathcal{D}_{\text{tr}}^{i}$
is sampled from the training task $\mathcal{T}_{\text{tr}}^{i} $.
As illustrated in Fig.~\ref{fig: maml pipeline}(A), this dataset is partitioned into a support set $\mathcal{D}_{\text{spt},\text{tr}}^{i} = \{(x_j, y_j)\}_{j=1}^{n}$ and a query set $\mathcal{D}_{\text{qry},\text{tr}}^{i} = \{(x_j, y_j)\}_{j=1}^{m}$, where $n$ and $m$ are the dataset sizes.

\subsubsection{Task-specific adaptation with support sets (Lines 5 and 6 in Algorithm~\ref{algorithm:meta-train})}
The purpose of task-specific adaptation is to simulate deployment-time fine-tuning, where the model must quickly specialize to new settings using only a small amount of calibration data. 

Given a model $f_{\theta}$ and a task-specific support set $\mathcal{D}_{\text{spt},\text{tr}}^{i}$,
the model parameters are adapted using gradient descent:
\begin{equation}
     \theta'_{i} =  \theta - \alpha\,\nabla_{ \theta}\,\mathcal{L}\left(f_{ \theta}; \mathcal{D}_{\text{spt},\text{tr}}^{i}\right),
    \label{eq:inner_loop_update}
\end{equation}
where $\alpha$ is a task-agnostic learning rate selected via hyperparameter tuning.

\subsubsection{Meta-update step using query sets (Lines 7 and 9 of Algorithm~\ref{algorithm:meta-train})}
The meta-update step 
begins by computing
the gradient of the loss for the adapted model $f_{\theta'_{i}}$
on its query set $\mathcal{D}_{\text{qry},\text{tr}}^{i}$.
The gradients $\nabla_\theta \mathcal{L}(f_{\theta'_{i}},\mathcal{D}_{\text{qry},\text{tr}}^{i})$ are aggregated across all tasks to update
$\theta$ as:
\begin{equation}
     \theta \leftarrow  \theta - \beta \nabla_{ \theta}\sum_{i}\mathcal{L}\left(f_{ \theta'_{i}};\mathcal{D}_{\text{qry},\text{tr}}^{i}\right)
    \label{eq:outer_loop_update}
\end{equation}
with $\beta$ a learning rate selected via hyperparameter tuning. 

Meta-update captures MAML's core idea: aggregating gradients across diverse tasks to improve the model initialization $\theta$. This preserves task-level variability during training, enabling rapid adaptation to unseen tasks with limited data.
In contrast, standard optimization methods such as stochastic gradient descent (SGD) update model parameters using gradients from mini-batches drawn from the overall data distribution, which may overlook task-specific structure and limit generalization to unseen settings.

\vspace{-0.15 in}
\subsection{Meta-testing Stage} \label{sec: meta_test_loop}
\vspace{-0.05 in}

Meta-testing 
evaluates the model’s ability to adapt to an unseen task
$\mathcal{T}_{\text{te}}$
by first fine-tuning the meta-trained model parameters $\theta$ on a support (calibration) set from $\mathcal{T}_{\text{te}}$, and then deploying the adapted model to make predictions on the task’s query set.
The unseen task may differ from training tasks, potentially introducing distribution shifts. For clarity, we focus on a single test task in this section.

As in meta-training, a support set is first sampled from $\mathcal{T}_{\text{te}}$, and model fine-tuning is performed using standard optimization algorithms such as SGD.
In our experiments, brief sensor recordings (2 to 5 seconds of walking data) are typically sufficient for calibration.
This high data efficiency is enabled by the adaptive initialization learned during meta-training, which facilitates rapid adaptation and accurate estimation on new tasks with minimal calibration and computation, as demonstrated in Sec.~\ref{sec: experiment}.

\begin{figure*}[ht]
    \centering
    \includegraphics[width= 0.98\textwidth]{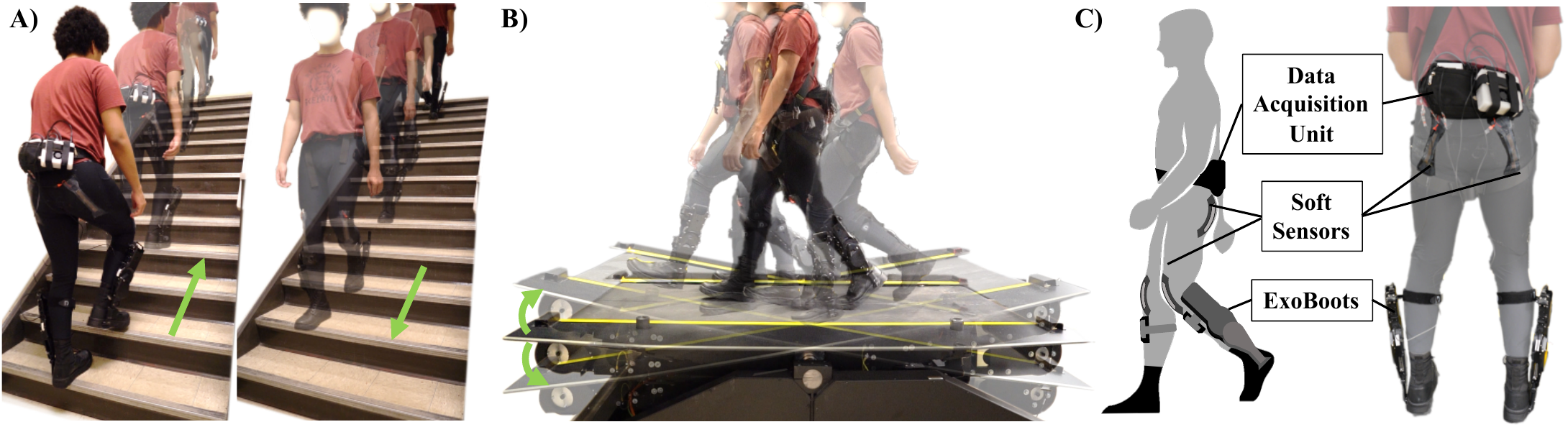}
    \vspace{-0.05 in}
    \caption{Experimental setup. (A) Out-of-lab stair trials, (B) in-lab inclined treadmill walking, and (C) soft sensors at the hip and knee, along with ExoBoots ankle exoskeletons used solely for ground-truth gait phase labeling during stair cases.}
    \label{fig: experiment condition}
\end{figure*}

\vspace{-0.1 in}
\section{Experiments} \label{sec: experiment}
\vspace{-0.05 in}

This section reports the experimental validation results, including comparison with various baseline methods.

\vspace{-0.15 in}
\subsection{Experimental setup}
\vspace{-0.05 in}

\subsubsection{Hardware}

The hardware setup (Fig.~\ref{fig: experiment condition}) includes an instrumented Motek M-Gait treadmill for configurable in-lab walking experiments and indoor stairs for out-of-lab trials.
In all experiments, subjects wore compression pants with four encapsulated fabric sensors~\cite{sanchez2023stretchable} embedded, as described in Sec.~\ref{sec: problem formulation},
as well as a pair of Dephy ExoBoots ankle exoskeletons.
Subjects also wore Dephy ExoBoots ankle exoskeletons used for ground-truth gait phase labeling during stair trials. To maintain consistent sensor placement and experimental conditions across treadmill and stair trials, subjects wore ExoBoots during treadmill experiments as well, although the ExoBoots data were not used. Model training and testing were conducted on a workstation with an Intel 14900K CPU and an Nvidia RTX 4090 GPU.

\subsubsection{Subjects}
To assess subject-level generalizability,
data were collected from nine participants (7 males, 2 females; age: $25 \pm 4$ years; height: $1.70 \pm 0.11$\,m; weight: $60 \pm 16$\,kg) under Purdue IRB approval (20-057-YAN-XPD). 

\subsubsection{Terrains and walking speeds}
For evaluation across different terrain types, walking experiments were conducted on a treadmill and a staircase (Fig.~\ref{fig: experiment condition}(A–B)).

Treadmill trials included \textbf{level walking (LW)}, \textbf{ramp ascent (RA)}, and \textbf{ramp descent (RD)} at incline angles of $0^\circ$, $\pm5^\circ$, and $\pm10^\circ$ .
Walking speeds were 0.9, 1.1, and 1.3\,m/s for LW, and 0.7, 0.9, and 1.1\,m/s for RA and RD.
Each subject completed two 2-minute trials per angle-speed combination on the same day.

\textbf{Stair ascent (SA)} and \textbf{stair descent (SD)} trials were conducted on a staircase with 18\,cm-high and 28\,cm-deep steps (approx. $33^\circ$ slope). Participants climbed 12 steps at self-selected speeds (approx. $0.8$-$1.0$\,m/s), completing five 10-second trials on a separate day from treadmill trials to reduce fatigue and ensure consistent performance. 

\subsubsection{Data acquisition} 
Raw capacitance signals from the soft sensors are acquired via 
a custom Arduino board and logged by a Nvidia Jetson Orin Nano.
The boards are carried in a fanny pack worn on the user’s lower back (Fig.~\ref{fig: experiment condition}(C)).

\subsubsection{Ground truth labeling}
Gait phases during treadmill trials are labeled using ground reaction forces, which are measured by the treadmill’s force plates at 1000\,Hz with a 5\,mm center-of-pressure error. 
For stair trials, subjects wear the ExoBoots, 
and a finite-state machine (FSM) detects heel-strike and toe-off events from the ExoBoots' accelerometer signals, with FSM thresholds calibrated against treadmill force plate data.
Ground-truth locomotion modes and terrain geometry are labeled based on the experimental protocol and pre-programmed treadmill inclines.

\vspace{-0.1 in}
\subsection{Baseline Setup}
\vspace{-0.05 in}

We compare the proposed MAML-based estimation framework against baseline learning strategies of increasing complexity to systematically evaluate adaptation efficiency and generalization across subjects and terrains.

For both the proposed and baseline approaches, the entire dataset is partitioned with eight subjects for training (if any)
and one held-out subject for testing (deployment).

\subsubsection{Baseline approaches}

All baseline models adopt the same DCNN architecture and weighted loss function as the proposed framework, using identical weights $w_{\text{gait}}$, $w_{\text{inc}}$, and $w_{\text{loc}}$. We evaluate the following baselines commonly used in meta-learning studies~\cite{finn2017metalearning,la2024meta} and in transfer-learning-based gait phase estimation~\cite{le2024transfer}:
\begin{enumerate}
    \item [(i)] \textbf{Random initialization (RI): no pre-training.}
    The DCNN model is trained from scratch on the support set of the unseen subject using randomly initialized parameters, without any pre-training on the other subjects.
    \item [(ii)] \textbf{Direct evaluation (DE): pre-training + no fine-tuning.} 
    The model is pre-trained on data from eight subjects and evaluated on the unseen subject without any parameter updates.
    \item [(iii)] \textbf{Transfer learning (TL): pre-training + partially freezing fine-tuning.}
    Following~\cite{le2024transfer}, the model is pre-trained on eight subjects and fine-tuned on the unseen subject, by updating only the top layers while earlier convolutional layers are frozen.
    \item [(iv)] \textbf{Standard fine-tuning (SFT): pre-training + full fine-tuning.}
    The model is pre-trained on the eight-subject dataset, and all layers are fine-tuned on the unseen test subject, reflecting a standard full fine-tuning approach.
    
\end{enumerate}

Note that the pre-training of DE, TL, and SFT is based on conventional supervised learning and does not employ the MAML pipeline.

\begin{figure*}[t]
    \centering
    \includegraphics[width= 0.99\textwidth]{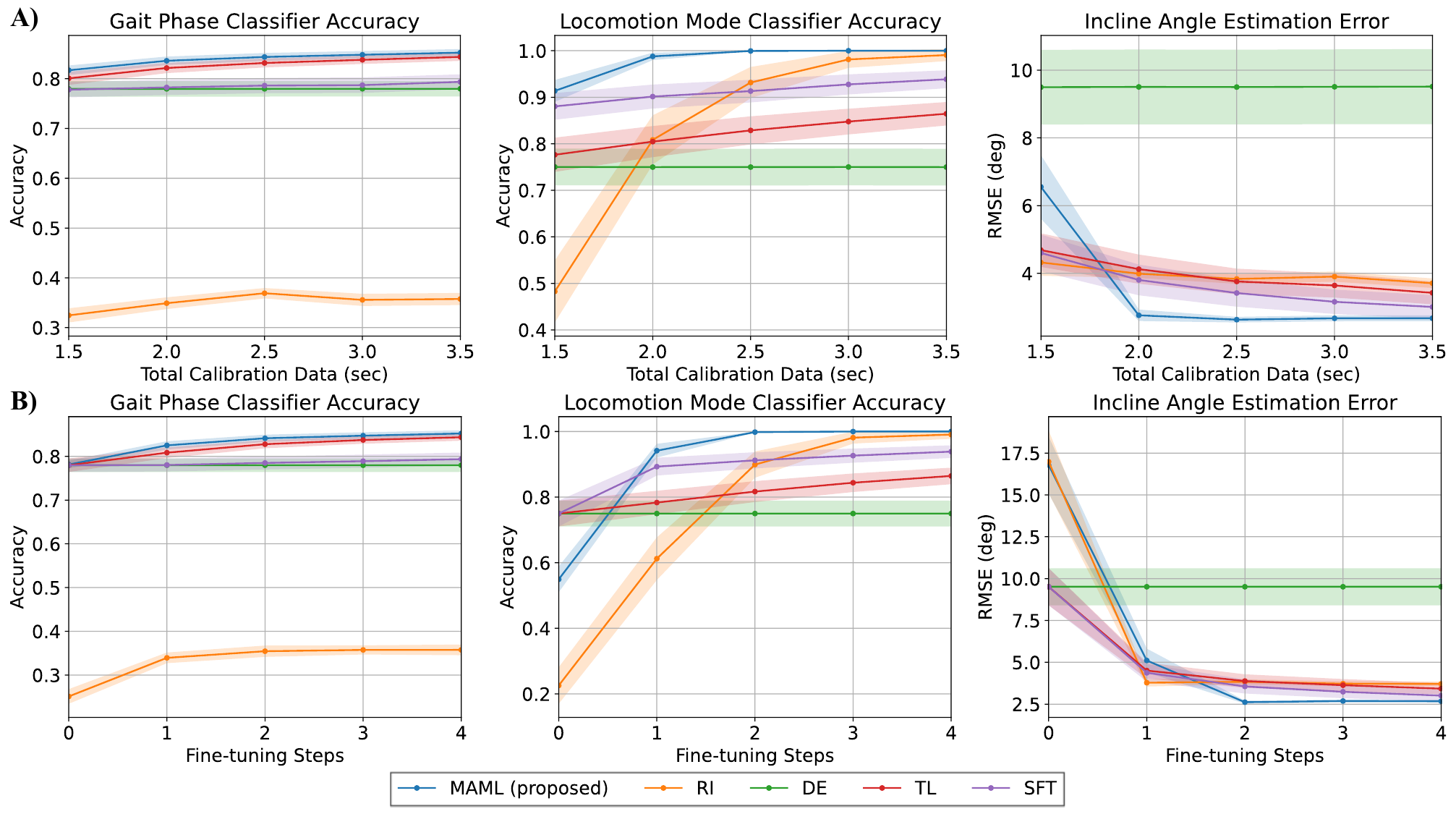}
    \vspace{-0.07 in}
    \caption{(A) Effect of calibration data size on estimation accuracy with four fine-tuning steps (scenario S1).
    (B) Effect of fine-tuning steps on estimation accuracy using 3.5\,s of calibration data (scenario S2).
    Solid lines and shaded areas denote mean and~95\% CI across subjects.}
    \label{fig: combined comparison plots}
    \vspace{0.3 cm }
\end{figure*}

\begin{figure*}[t]
    \centering
    \includegraphics[width= 1\textwidth]{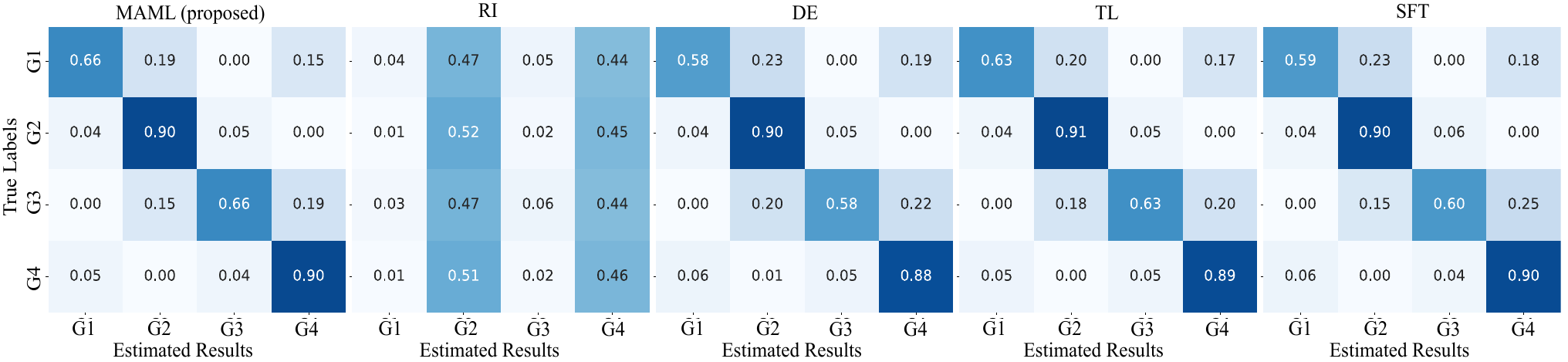}
    \vspace{-0.07 in}
    \caption{Confusion matrices for gait phase classification using 3.5\,s of calibration data and two fine-tune steps (scenario S2). Darker diagonal cells indicate higher accuracy.}
    \label{fig: confusion matrices}
    \vspace{0.2 cm}
\end{figure*}

\vspace{-0.1 in}
\subsection{Training Setup}
\vspace{-0.05 in}

\subsubsection{Loss function weight}
The same loss weights are used for both MAML and baseline models during training and fine-tuning.
We assign $w_{\text{gait}}=0.6$ and $w_{\text{inc}}=w_{\text{loc}}=0.2$. 
The locomotion mode loss is down-weighted due to its indirect optimization through gait and incline estimation, while the incline angle loss is reduced because its regression head converges quickly during training. 

\subsubsection{MAML training}
For meta-training,
each task dataset is divided into a support set of 80 samples and a query set of 120 samples ($n=80$ and $m=120$)
to balance adaptation and meta-update. 
Meta‐training is performed for 200 epochs (each epoch iterates over all tasks in Algorithm~\ref{algorithm:meta-train}) to ensure convergence of model parameters.

\subsubsection{Baseline training}

The baselines DE, TL, and SFT are trained using the Adam optimizer~\cite{kingma2014adam} with 200 data points per batch. 
Training is conducted on the combined dataset from eight subjects until convergence, using the same input-output format $(x, y)$ as in MAML.
Since RI does not include a pre-training stage, it is excluded from this process.

\vspace{-0.1 in}
\subsection{Testing Setup}
\vspace{-0.05 in}

\subsubsection{Hyperparameter selection for fine-tuning} 
To ensure fair comparison of fine-tuning performance, we apply Bayesian optimization~\cite{BayesianOptimization} to select the optimal SGD learning rate for each method, aiming to maximize average prediction accuracy after fine-tuning. The selected learning rates are: $3\times10^{-4}$ for MAML; $1.5\times10^{-4}$ for RI; $8\times10^{-4}$ for TL; and $1\times10^{-6}$ for SFT. 
DE is excluded from this process, as it does not involve fine-tuning.

\subsubsection{Evaluation scenarios for adaptation and generalization}
To evaluate each approach's adaptation to an unseen subject under varying calibration data sizes and fine-tuning steps, as well as its post-adaptation generalization across subjects and locomotion modes, we consider three scenarios:
\textbf{(S1) Data efficiency of adaptation:} 
Fixing four fine-tuning steps, we vary calibration data duration (across 1.5, 2, 2.5, 3, and 3.5\,s per locomotion mode and speed) to assess adaptation with brief data.
\textbf{(S2) Computation efficiency of adaptation.} 
Fixing calibration data to 3.5\,s, we vary the number of fine-tuning steps (0–4) to evaluate adaptation speed.
\textbf{(S3) Generalizability across subjects and locomotion modes.}  
With both calibration data (3.5\,s) and fine-tuning steps (4) fixed, we compare estimation accuracy across individual subjects and locomotion modes to assess post-adaptation generalization.

\subsubsection{Estimation accuracy measures}
Accuracy for gait phase and locomotion mode classification is computed as the total number of correct predictions divided by the total number of test query samples. 
Incline angle accuracy is evaluated using RMSE. All results are presented as mean~$\pm$~95\% confidence interval (CI).

\begin{table*}[t]
\centering
\caption{\centering 
Subject-specific estimation accuracy using 3.5\,s of calibration data and four fine-tuning steps, i.e., under Scenario (S3). Values show mean accuracy (classification) or RMSE (incline angle) with a 95\% CI.
The table highlights \textbf{the best performance}, \underline{the second-best performance}, and \colorbox{lightblue}{proposed method}.}
\vspace{-0.05 in}
\resizebox{\textwidth}{!}{%
\begin{tabular}{cccccccccccc}
\toprule
 & \textbf{Method} & \textbf{Sub01} & \textbf{Sub02} & \textbf{Sub03} & \textbf{Sub04} & \textbf{Sub05} & \textbf{Sub06} & \textbf{Sub07} & \textbf{Sub08} & \textbf{Sub09} & \textbf{Average} \\
\midrule
\multirow{5}{*}{\rotatebox[origin=c]{90}{\makecell{Gait Phase \\ Accuracy}}}
 & \cellcolor{lightblue}MAML & \cellcolor{lightblue}$\mathbf{0.84_{\pm 0.02}}$ & \cellcolor{lightblue}$\underline{0.86_{\pm 0.02}}$ & \cellcolor{lightblue}$\underline{0.85_{\pm 0.03}}$ & \cellcolor{lightblue}$\mathbf{0.88_{\pm 0.01}}$ & \cellcolor{lightblue}$\mathbf{0.87_{\pm 0.01}}$ & \cellcolor{lightblue}$\underline{0.84_{\pm 0.02}}$ & \cellcolor{lightblue}$\mathbf{0.86_{\pm 0.02}}$ & \cellcolor{lightblue}$\mathbf{0.84_{\pm 0.02}}$ &  \cellcolor{lightblue}$\mathbf{0.82_{\pm 0.03}}$ &\cellcolor{lightblue}$\mathbf{0.85_{\pm 0.01}}$ \\
 & RI & $0.36_{\pm 0.03}$ & $0.34_{\pm 0.04}$ & $0.36_{\pm 0.03}$ & $0.35_{\pm 0.04}$ & $0.36_{\pm 0.04}$ & $0.35_{\pm 0.05}$ & $0.39_{\pm 0.04}$ & $0.35_{\pm 0.03}$ & $0.36_{\pm 0.04}$ & $0.36_{\pm 0.01}$ \\
 & DE & $0.75_{\pm 0.06}$ & $0.85_{\pm 0.02}$ & $0.79_{\pm 0.05}$ & $0.78_{\pm 0.03}$ & $0.81_{\pm 0.02}$ & $0.79_{\pm 0.03}$ & $0.80_{\pm 0.03}$ & $0.78_{\pm 0.03}$ & $0.68_{\pm 0.07}$ & $0.78_{\pm 0.02}$\\
 & TL & $\underline{0.83_{\pm 0.03}}$ & $\mathbf{0.88_{\pm 0.02}}$ & $\mathbf{0.86_{\pm 0.02}}$ & $\underline{0.86_{\pm 0.02}}$ & $\underline{0.85_{\pm 0.01}}$ & $\mathbf{0.85_{\pm 0.02}}$ & $\underline{0.85_{\pm 0.02}}$ & $\underline{0.83_{\pm 0.02}}$ & $\underline{0.80_{\pm 0.03}}$ & $\underline{0.84_{\pm 0.01}}$\\
 & SFT & $0.79_{\pm 0.06}$ & $\underline{0.86_{\pm 0.02}}$ & $0.79_{\pm 0.05}$ & $0.78_{\pm 0.04}$ & $0.83_{\pm 0.03}$ & $0.79_{\pm 0.03}$ & $0.82_{\pm 0.02}$ & $0.80_{\pm 0.03}$ & $0.69_{\pm 0.07}$ & $0.79_{\pm 0.02}$ \\
\midrule
\multirow{5}{*}{\rotatebox[origin=c]{90}{\makecell{Loc. Mode \\Accuracy}}}
 & \cellcolor{lightblue}MAML & \cellcolor{lightblue}$\mathbf{1.00_{\pm 0.00}}$ & \cellcolor{lightblue}$\mathbf{1.00_{\pm 0.00}}$ & \cellcolor{lightblue}$\mathbf{1.00_{\pm 0.00}}$ & \cellcolor{lightblue}$\mathbf{1.00_{\pm 0.00}}$ & \cellcolor{lightblue}$\mathbf{1.00_{\pm 0.00}}$ & \cellcolor{lightblue}$\mathbf{1.00_{\pm 0.00}}$ & \cellcolor{lightblue}$\mathbf{1.00_{\pm 0.00}}$ & \cellcolor{lightblue}$\mathbf{1.00_{\pm 0.00}}$ & \cellcolor{lightblue}$\mathbf{1.00_{\pm 0.00}}$ & \cellcolor{lightblue}$\mathbf{1.00_{\pm 0.00}}$\\
 & RI & $\mathbf{1.00_{\pm 0.00}}$ & $\mathbf{1.00_{\pm 0.00}}$ & $\mathbf{1.00_{\pm 0.00}}$ & $\underline{0.96_{\pm 0.01}}$ & $\underline{0.96_{\pm 0.08}}$ & $\mathbf{1.00_{\pm 0.00}}$ & $\mathbf{1.00_{\pm 0.00}}$ & $\mathbf{1.00_{\pm 0.00}}$ & $\mathbf{1.00_{\pm 0.00}}$ & $\underline{0.99_{\pm 0.01}}$ \\
 & DE & $0.75_{\pm 0.14}$ & $0.83_{\pm 0.11}$ & $0.72_{\pm 0.15}$ & $0.80_{\pm 0.07}$ & $0.81_{\pm 0.08}$ & $0.80_{\pm 0.09}$ & $0.91_{\pm 0.03}$ & $0.76_{\pm 0.07}$ &  $0.38_{\pm 0.14}$ &  $0.75_{\pm 0.04}$\\
 & TL & $0.91_{\pm 0.07}$ & $0.88_{\pm 0.08}$ & $0.83_{\pm 0.11}$ & $0.91_{\pm 0.04}$ & $0.87_{\pm 0.06}$ & $0.90_{\pm 0.06}$ & $0.95_{\pm 0.02}$ & $0.89_{\pm 0.05}$ & $0.65_{\pm 0.10}$ &  $0.87_{\pm 0.03}$\\
 & SFT & $0.87_{\pm 0.10}$ & $0.98_{\pm 0.02}$ & $0.95_{\pm 0.05}$ & $0.94_{\pm 0.05}$ & $0.98_{\pm 0.01}$ & $0.98_{\pm 0.02}$ & $0.98_{\pm 0.01}$ & $0.95_{\pm 0.03}$ & $0.82_{\pm 0.10}$ &  $0.94_{\pm 0.02}$\\
\midrule
\multirow{5}{*}{\rotatebox[origin=c]{90}{\makecell{Incline \\ RMSE ($^\circ$)}}}
 & \cellcolor{lightblue}MAML & \cellcolor{lightblue}$\underline{2.75_{\pm 0.30}}$ & \cellcolor{lightblue}$\underline{2.57_{\pm 0.22}}$ & \cellcolor{lightblue}$\underline{2.51_{\pm 0.15}}$ & \cellcolor{lightblue}$\mathbf{2.63_{\pm 0.17}}$ & \cellcolor{lightblue}$\underline{2.69_{\pm 0.20}}$ & \cellcolor{lightblue}$\mathbf{2.60_{\pm 0.19}}$ & \cellcolor{lightblue}$\mathbf{2.54_{\pm 0.18}}$ & \cellcolor{lightblue}$\underline{2.79_{\pm 0.28}}$ & \cellcolor{lightblue}$\underline{2.98_{\pm 0.41}}$ & \cellcolor{lightblue}$\mathbf{2.67_{\pm 0.09}}$\\
 & RI & $3.56_{\pm 0.40}$ & $3.67_{\pm 0.48}$ & $3.89_{\pm 0.41}$ & $3.73_{\pm 0.38}$ & $3.60_{\pm 0.27}$ & $3.81_{\pm 0.51}$ & $\underline{3.81_{\pm 0.50}}$ & $3.53_{\pm 0.36}$ & $3.77_{\pm 0.40}$ & $3.71_{\pm 0.14}$ \\
 & DE & $7.52_{\pm 3.35}$ & $5.19_{\pm 1.29}$ & $9.18_{\pm 1.92}$ & $13.94_{\pm 4.73}$ & $6.97_{\pm 1.83}$ & $11.66_{\pm 3.61}$ & $9.94_{\pm 3.48}$ & $7.82_{\pm 2.05}$ & $12.26_{\pm 3.68}$ & $9.51_{\pm 1.11}$\\
 & TL & $3.74_{\pm 1.69}$ & $2.57_{\pm 0.72}$ & $3.10_{\pm 0.55}$ & $\underline{3.46_{\pm 0.74}}$ & $2.86_{\pm 0.49}$ & $3.95_{\pm 1.13}$ & $4.10_{\pm 1.49}$ & $3.36_{\pm 0.67}$ & $3.54_{\pm 0.83}$ & $3.42_{\pm 0.34}$ \\
 & SFT & $\mathbf{2.40_{\pm 0.65}}$ & $\mathbf{1.89_{\pm 0.34}}$ & $\mathbf{2.47_{\pm 0.39}}$ & $4.96_{\pm 1.75}$ & $\mathbf{2.60_{\pm 0.62}}$ & $\underline{3.13_{\pm 0.68}}$ & $4.21_{\pm 1.73}$ & $\mathbf{2.46_{\pm 0.40}}$ & $\mathbf{2.63_{\pm 0.47}}$  & $\underline{3.01_{\pm 0.35}}$ \\
\bottomrule
\end{tabular}
}%
\label{tab: subject results}
\vspace{+0.1 in}
\end{table*}

\vspace{-0.15 in}
\subsection{Result Analysis and Discussion} \label{sec: result discussion}
\vspace{-0.05 in}

We compare the performance of all approaches as follows.

\subsubsection{Comparison on data efficiency of adaptation} \label{sec: varied spt size}

Using the test results from Scenario (S1), we evaluate the calibration data efficiency of the proposed MAML-based estimation framework and baseline methods under varying calibration data sizes during deployment. Figure\,\ref{fig: combined comparison plots}(A) presents the estimation accuracy of all algorithms with a fixed number of fine-tuning steps across different calibration data durations.

{\textbf{MAML achieves the highest calibration data efficiency in adaptation.}}
As calibration data increases to 2\,s and beyond, the proposed MAML-based estimation framework consistently reaches the highest accuracy for all three estimation problems among all approaches. This confirms its ability to adapt effectively to unseen subjects using limited data. In comparison, TL and SFT require significantly more data to reach similar accuracy.

{\textbf{Pre-training is critical for accurate gait phase classification.}}
RI underperforms significantly in gait phase classification compared to terrain geometry and locomotion mode estimation. This is likely due to the fine-grained, high-temporal-resolution features needed for phase estimation, which are difficult to learn from scratch with limited data. In contrast, locomotion mode classification involves more coarse-grained features and is less sensitive to pre-training, explaining RI’s comparatively stronger performance in locomotion mode classification compared to gait phase estimation.

{\it {\bf Standard pre-training is not always superior to training from scratch, especially with sufficient task-specific calibration data.}}
With sufficient calibration data (e.g., $>$2.5\,s), RI surpasses all methods that rely on pre-training, except MAML, in locomotion mode classification.
Also, RI achieves similar accuracy as TL and SFT in incline angle estimation.
This suggests that with sufficient task-specific calibration data available, standard pre-training offers limited benefit for some estimation problems.

\subsubsection{Comparison on computation efficiency of adaptation} \label{sec: varied fine tune}

Figure~\ref{fig: combined comparison plots}(B) displays
the estimation accuracy results under Scenario (S2) where the computational efficiency of fine-tuning is evaluated by using a fixed 3.5\,s of calibration data and varying the number of fine-tuning steps from 0 to 4. 
The accuracy at zero fine-tuning steps reflects the zero-shot generalization capability of each method. Notably, SFT and TL without fine-tuning reduce to DE. Since DE involves no fine-tuning, its performance remains constant across all conditions and serves as the baseline for zero-shot accuracy.

{\bf MAML enables computationally efficient fine-tuning.}
Figure~\ref{fig: combined comparison plots}(B) shows that while MAML matches DE in gait phase classification at zero finte-tuning steps, it performs slightly worse in locomotion mode and incline angle estimation. However, with just one or two fine-tuning steps, MAML rapidly outperforms all baseline methods across all three estimation problems. This highlights MAML’s strength in computational adaptation efficiency: instead of optimizing solely for zero-shot generalization, it learns model parameters that are highly adaptable to new conditions, enabling accurate estimation with minimal fine-tuning.

\subsubsection{Analysis of gait phase misclassification} \label{sec: confusion matrix}

To provide general insights into gait phase classification,
Fig.~\ref{fig: confusion matrices} presents the confusion matrices for each method after two fine-tuning steps using 3.5\,s of support data, which is a special case under Scenario (S2). 

\textbf{Most gait phase misclassifications tend to occur between adjacent phases across all methods}, as reflected by the off-diagonal entries.
These errors may be partially attributed to the inherent ambiguity near phase transitions, as well as the inevitable labeling inaccuracies caused by hardware imperfections (e.g., noise and biases in force-plate and accelerometer data).

\subsubsection{Comparison on generalization across subjects and locomotion modes} \label{sec: vaired task}

\begin{table}[t]
\centering
\caption{\centering Locomotion-mode-specific estimation accuracy using 3.5\,s of calibration data and four fine-tuning steps, i.e., under Scenario (S3).
Formatting follows Table~\ref{tab: subject results}.}
\vspace{-0.05 in}
\resizebox{0.48\textwidth}{!}{%
\begin{tabular}{cccccccc}
\toprule
 & \textbf{Method} & \textbf{LW} & \textbf{RA} & \textbf{RD} & \textbf{SA} & \textbf{SD}\\
\midrule

\multirow{5}{*}{\rotatebox[origin=c]{90}{\makecell{Gait Phase \\ Accuracy}}}
 & \cellcolor{lightblue}MAML & \cellcolor{lightblue}$\mathbf{0.89_{\pm 0.01}}$ & \cellcolor{lightblue}\underline{$0.85_{\pm 0.01}$} & \cellcolor{lightblue}$\mathbf{0.85_{\pm 0.01}}$ &  \cellcolor{lightblue}$\mathbf{0.85_{\pm 0.02}}$ & \cellcolor{lightblue}$\mathbf{0.84_{\pm 0.02}}$ \\
 & RI & $0.40_{\pm 0.01}$ & $0.38_{\pm 0.01}$ & $0.40_{\pm 0.00}$ & $0.28_{\pm 0.03}$ & $0.32_{\pm 0.03}$ \\
 & DE & $0.87_{\pm 0.02}$ & $0.82_{\pm 0.02}$ & $0.82_{\pm 0.02}$ & $0.73_{\pm 0.04}$ & $0.66_{\pm 0.04}$ \\
 & TL & \underline{$0.88_{\pm 0.01}$} & \underline{$0.85_{\pm 0.01}$} & \underline{$0.84_{\pm 0.01}$} & $\mathbf{0.85_{\pm 0.02}}$ & \underline{$0.81_{\pm 0.02}$} \\
 & SFT & $0.87_{\pm 0.02}$ & $\mathbf{0.86_{\pm 0.01}}$ & \underline{$0.84_{\pm 0.01}$} & $0.72_{\pm 0.04}$ & $0.67_{\pm 0.04}$ \\
\midrule

\multirow{5}{*}{\rotatebox[origin=c]{90}{\makecell{Loc. Mode\\ Accuracy}}}
 & \cellcolor{lightblue}MAML & \cellcolor{lightblue}$\mathbf{1.00_{\pm 0.00}}$ & \cellcolor{lightblue}$\mathbf{1.00_{\pm 0.00}}$ & \cellcolor{lightblue}$\mathbf{1.00_{\pm 0.00}}$ & \cellcolor{lightblue}$\mathbf{1.00_{\pm 0.00}}$ & \cellcolor{lightblue}$\mathbf{1.00_{\pm 0.00}}$ \\
 & RI & $\mathbf{1.00_{\pm 0.00}}$ & $\mathbf{1.00_{\pm 0.00}}$ & $\mathbf{1.00_{\pm 0.00}}$ & \underline{$0.97_{\pm 0.05}$} & $\mathbf{1.00_{\pm 0.00}}$ \\
 & DE & $0.49_{\pm 0.11}$ & $0.95_{\pm 0.02}$ & $0.62_{\pm 0.08}$ & $0.76_{\pm 0.09}$ & $0.82_{\pm 0.09}$  \\
 & TL & $0.70_{\pm 0.08}$ & $0.97_{\pm 0.01}$ & $0.76_{\pm 0.05}$ & $0.88_{\pm 0.06}$ & $0.97_{\pm 0.02}$ \\
 & SFT & $0.91_{\pm 0.05}$ & $0.99_{\pm 0.01}$ & $0.97_{\pm 0.02}$ & $0.81_{\pm 0.08}$ & $0.98_{\pm 0.02}$ \\
\midrule

\multirow{5}{*}{\rotatebox[origin=c]{90}{\makecell{Incline \\ RMSE ($^\circ$)}}}
 & \cellcolor{lightblue}MAML & \cellcolor{lightblue}$2.18_{\pm 0.05}$ & \cellcolor{lightblue}$2.32_{\pm 0.05}$ & \cellcolor{lightblue}\underline{$2.31_{\pm 0.04}$} & \cellcolor{lightblue}$\mathbf{3.19_{\pm 0.13}}$ & \cellcolor{lightblue}$\mathbf{3.39_{\pm 0.21}}$ \\
 & RI & $3.45_{\pm 0.31}$ & $3.61_{\pm 0.22}$ & $3.73_{\pm 0.24}$ & \underline{$4.24_{\pm 0.26}$} & \underline{$4.14_{\pm 0.25}$} \\
 & DE & $3.89_{\pm 0.61}$ & $3.55_{\pm 0.38}$ & $5.64_{\pm 0.66}$ & $17.79_{\pm 2.06}$ & $18.47_{\pm 2.61}$ \\
 & TL & \underline{$2.12_{\pm 0.17}$} & \underline{$1.89_{\pm 0.11}$} & $2.37_{\pm 0.15}$ & $6.03_{\pm 1.05}$ & $5.09_{\pm 0.70}$ \\
 & SFT & $\mathbf{1.54_{\pm 0.11}}$ & $\mathbf{1.51_{\pm 0.07}}$ & $\mathbf{1.83_{\pm 0.08}}$ & $5.68_{\pm 1.19}$ & $4.69_{\pm 0.42}$ \\
\bottomrule
\end{tabular}
}%
\label{tab: locmode results}
\vspace{-0.2 in}
\end{table}

Table~\ref{tab: subject results} reports estimation accuracy for each unseen subject in a leave-one-subject-out cross-validation setup, where the model is trained on data from the remaining eight subjects. Table~\ref{tab: locmode results} presents the performance for each locomotion mode, 
aggregated over all nine held-out test subjects.
All results are obtained under Scenario (S3), which uses 3.5\,s of calibration data and four fine-tuning steps for adaptation.
These per-subject and per-mode breakdowns reveal variations in estimation performance not captured by aggregate metrics shown earlier.

{\bf Each estimation approach shows strengths under specific conditions.} 
From Tables~\ref{tab: subject results} and \ref{tab: locmode results}, MAML consistently outperforms all baselines on average, across subjects and locomotion modes.
Also, each baseline shows strengths suited to particular aspects of estimation:
(a) TL performs comparably to MAML in gait phase classification across all subjects and locomotion modes. suggests that its frozen feature extractor captures generalizable temporal gait patterns, while its trainable head adapts to subject-specific nuances.
(b) SFT yields the lowest incline RMSE for most subjects (except Sub04, Sub06, and Sub07) and for most locomotion modes (except SA and SD), indicating that full-network fine-tuning is advantageous for capturing subtle, terrain-specific features important to incline estimation.
(c) RI, though generally less accurate, performs strongly in locomotion mode classification,
which is consistent with findings in Sec.~\ref{sec: varied spt size} that training from scratch is effective for this high-level categorical estimation problem under sufficient calibration data.

{\bf MAML mitigates the effects of data imbalance.} 
In our experiments, treadmill walking trials provide more data than stair climbing trials, resulting in data imbalance. Likely due to this imbalance, DE, TL, and SFT, which rely on standard supervised learning, exhibit elevated errors in stair scenarios (SA and SD) compared to treadmill walking (LW, RA, and RD). In contrast, MAML maintains robust performance across all locomotion modes, benefiting from its task-specific adaptation and gradient aggregation across diverse conditions.

\vspace{-0.15 in}
\section{Conclusion} \label{sec: conclusion}
\vspace{-0.05 in}

This paper introduced a MAML-based estimation approach using fabric-based soft wearable sensors for simultaneous gait phase and terrain geometry estimation.
The estimation problem was formulated as a combination of classification (for gait phase and locomotion mode estimation) and regression (for incline angle estimation).
The proposed DCNN architecture with a multi-head output structure, combined with MAML to capture subject- and terrain-invariant structure, enables accurate, efficient adaptation to new users and robust generalization across subjects and terrains.
Experiments across nine subjects and five terrains confirmed that our approach consistently outperforms conventional supervised and transfer learning baselines in estimation accuracy, adaptation efficiency, and generalization.

\vspace{-0.05 in}
\bibliographystyle{IEEEtran}
\bibliography{Reference.bib}

\begin{thebibliography}{10}
\providecommand{\url}[1]{#1}
\csname url@samestyle\endcsname
\providecommand{\newblock}{\relax}
\providecommand{\bibinfo}[2]{#2}
\providecommand{\BIBentrySTDinterwordspacing}{\spaceskip=0pt\relax}
\providecommand{\BIBentryALTinterwordstretchfactor}{4}
\providecommand{\BIBentryALTinterwordspacing}{\spaceskip=\fontdimen2\font plus
\BIBentryALTinterwordstretchfactor\fontdimen3\font minus \fontdimen4\font\relax}
\providecommand{\BIBforeignlanguage}[2]{{%
\expandafter\ifx\csname l@#1\endcsname\relax
\typeout{** WARNING: IEEEtran.bst: No hyphenation pattern has been}%
\typeout{** loaded for the language `#1'. Using the pattern for}%
\typeout{** the default language instead.}%
\else
\language=\csname l@#1\endcsname
\fi
#2}}
\providecommand{\BIBdecl}{\relax}
\BIBdecl

\bibitem{young2016state}
A.~J. Young and D.~P. Ferris, ``State of the art and future directions for lower limb robotic exoskeletons,'' \emph{IEEE Trans. Neural Syst. Rehabil. Eng.}, vol.~25, no.~2, pp. 171--182, 2016.

\bibitem{cortino2023data}
R.~J. Cortino, T.~K. Best, and R.~D. Gregg, ``Data-driven phase-based control of a powered knee-ankle prosthesis for variable-incline stair ascent and descent,'' \emph{IEEE Trans. Med. Robot. Bionics.}, vol.~6, no.~1, pp. 175--188, 2023.

\bibitem{chinimilli2019automatic}
P.~T. Chinimilli, Z.~Qiao, S.~M.~R. Sorkhabadi, V.~Jhawar, I.~H. Fong, and W.~Zhang, ``Automatic virtual impedance adaptation of a knee exoskeleton for personalized walking assistance,'' \emph{Robot. Auton. Syst.}, vol. 114, pp. 66--76, 2019.

\bibitem{cheng2025ambilateral}
S.~Cheng, C.~A. Laubscher, T.~K. Best, and R.~D. Gregg, ``Ambilateral activity recognition and continuous adaptation with a powered knee-ankle prosthesis,'' \emph{IEEE Trans. Robot.}, 2025.

\bibitem{alzubaidi2021review}
L.~Alzubaidi, J.~Zhang, A.~J. Humaidi, A.~Al-Dujaili, Y.~Duan, O.~Al-Shamma, J.~Santamar{\'\i}a, M.~A. Fadhel, M.~Al-Amidie, and L.~Farhan, ``Review of deep learning: concepts, cnn architectures, challenges, applications, future directions,'' \emph{J. Big Data}, vol.~8, no.~1, p.~53, 2021.

\bibitem{hou2020pedestrian}
X.~Hou and J.~Bergmann, ``Pedestrian dead reckoning with wearable sensors: A systematic review,'' \emph{IEEE Sens. J.}, vol.~21, no.~1, pp. 143--152, 2020.

\bibitem{yuan20133}
Q.~Yuan and I.-M. Chen, ``3-d localization of human based on an inertial capture system,'' \emph{IEEE Trans. Robot.}, vol.~29, no.~3, pp. 806--812, 2013.

\bibitem{zhang2015whole}
Y.~Zhang, K.~Chen, J.~Yi, T.~Liu, and Q.~Pan, ``Whole-body pose estimation in human bicycle riding using a small set of wearable sensors,'' \emph{IEEE/ASME Trans. Mechatron.}, vol.~21, no.~1, pp. 163--174, 2015.

\bibitem{kim2018deep}
D.~Kim, J.~Kwon, S.~Han, Y.-L. Park, and S.~Jo, ``Deep full-body motion network for a soft wearable motion sensing suit,'' \emph{IEEE/ASME Trans. Mechatron.}, vol.~24, no.~1, pp. 56--66, 2018.

\bibitem{li2023shortcut}
X.~Li, C.~Ye, B.~Huang, Z.~Zhou, Y.~Su, Y.~Ma, Z.~Yi, and X.~Wu, ``A shortcut enhanced lstm-gcn network for multi-sensor based human motion tracking,'' \emph{IEEE Trans. Autom. Sci. Eng.}, 2023.

\bibitem{molinaro2024task}
D.~D. Molinaro, K.~L. Scherpereel, E.~B. Schonhaut, G.~Evangelopoulos, M.~K. Shepherd, and A.~J. Young, ``Task-agnostic exoskeleton control via biological joint moment estimation,'' \emph{Nature}, vol. 635, no. 8038, pp. 337--344, 2024.

\bibitem{xu2025reciprocal}
J.~Xu, A.~Chen, L.~Winterbottom, J.~Palacios, P.~Chivukula, D.~M. Nilsen, J.~Stein, and M.~Ciocarlie, ``Reciprocal learning of intent inferral with augmented visual feedback for stroke,'' in \emph{2025 International Conference On Rehabilitation Robotics (ICORR)}.\hskip 1em plus 0.5em minus 0.4em\relax IEEE, 2025, pp. 1512--1517.

\bibitem{kowalsky2021human}
D.~B. Kowalsky, J.~R. Rebula, L.~V. Ojeda, P.~G. Adamczyk, and A.~D. Kuo, ``Human walking in the real world: Interactions between terrain type, gait parameters, and energy expenditure,'' \emph{PLoS one}, vol.~16, no.~1, p. e0228682, 2021.

\bibitem{xu2024chatemg}
J.~Xu, R.~Wang, S.~Shang, A.~Chen, L.~Winterbottom, T.-L. Hsu, W.~Chen, K.~Ahmed, P.~L.~L. Rotta, X.~Zhu, D.~M. Nilsen, J.~Stein, and M.~Ciocarlie, ``Chatemg: Synthetic data generation to control a robotic hand orthosis for stroke,'' \emph{IEEE Robotics and Automation Letters}, vol.~9, no.~2, pp. 1--8, 2024.

\bibitem{le2024transfer}
D.~Le, S.~Cheng, R.~D. Gregg, and M.~Ghaffari, ``Transfer learning for efficient intent prediction in lower-limb prosthetics: A strategy for limited datasets,'' \emph{IEEE Robot. Autom. Lett.}, 2024.

\bibitem{hospedales2021meta}
T.~Hospedales, A.~Antoniou, P.~Micaelli, and A.~Storkey, ``Meta-learning in neural networks: A survey,'' \emph{IEEE Trans. Pattern Anal. Mach. Intell.}, vol.~44, no.~9, pp. 5149--5169, 2021.

\bibitem{cao2024fusion}
W.~Cao, C.~Li, L.~Yang, M.~Yin, C.~Chen, W.~Kobsiriphat, T.~Utakapan, Y.~Yang, H.~Yu, and X.~Wu, ``A fusion network with stacked denoise autoencoder and meta learning for lateral walking gait phase recognition and multi-step-ahead prediction,'' \emph{IEEE J. Biomed. Health Inform.}, 2024.

\bibitem{la2024meta}
P.~L. La~Rotta, J.~Xu, A.~Chen, L.~Winterbottom, W.~Chen, D.~Nilsen, J.~Stein, and M.~Ciocarlie, ``Meta-learning for fast adaptation in intent inferral on a robotic hand orthosis for stroke,'' in \emph{Proc. IEEE/RSJ Int. Conf. Intell. Robots Syst.}\hskip 1em plus 0.5em minus 0.4em\relax IEEE, 2024, pp. 4693--4700.

\bibitem{zhu2022design}
Z.~Zhu, S.~M.~R. Sorkhabadi, Y.~Gu, and W.~Zhang, ``Design and evaluation of an invariant extended kalman filter for trunk motion estimation with sensor misalignment,'' \emph{IEEE/ASME Trans. Mechatron.}, vol.~27, no.~4, pp. 2158--2167, 2022.

\bibitem{leestma2023linking}
J.~K. Leestma, P.~R. Golyski, C.~R. Smith, G.~S. Sawicki, and A.~J. Young, ``Linking whole-body angular momentum and step placement during perturbed human walking,'' \emph{J. Exp. Biol.}, vol. 226, no.~6, p. jeb244760, 2023.

\bibitem{Walshwearablesensor}
Y.~Mengüç, Y.-L. Park, H.~Pei, D.~Vogt, P.~M. Aubin, E.~Winchell, L.~Fluke, L.~Stirling, R.~J. Wood, and C.~J. Walsh, ``Wearable soft sensing suit for human gait measurement,'' \emph{Int. J. Robot. Res.}, vol.~33, no.~14, pp. 1748--1764, 2014.

\bibitem{sanchez2023stretchable}
L.~Sanchez-Botero, A.~Agrawala, and R.~Kramer-Bottiglio, ``Stretchable, breathable, and washable fabric sensor for human motion monitoring,'' \emph{Adv. Mater. Tech.}, p. 2300378, 2023.

\bibitem{case2019robotic}
J.~C. Case, M.~C. Yuen, J.~Jacobs, and R.~Kramer-Bottiglio, ``Robotic skins that learn to control passive structures,'' \emph{IEEE Robot. Autom. Lett.}, vol.~4, no.~3, pp. 2485--2492, 2019.

\bibitem{zhang2023intelligent}
W.~Zhang, A.~Tashakori, Z.~Jiang, A.~Servati, H.~Narayana, S.~Soltanian, R.~Y. Yeap, M.~Ma, L.~Toy, and P.~Servati, ``Intelligent knee sleeves: A real-time multimodal dataset for 3d lower body motion estimation using smart textile,'' \emph{Adv. Neural Inf. Process. Syst.}, vol.~36, pp. 42\,502--42\,515, 2023.

\bibitem{kharb2011review}
A.~Kharb, V.~Saini, Y.~Jain, and S.~Dhiman, ``A review of gait cycle and its parameters,'' \emph{Int. J. Comput. Eng. Manag.}, vol.~13, no.~01, pp. 78--83, 2011.

\bibitem{finn2017metalearning}
C.~Finn, P.~Abbeel, and S.~Levine, ``Model-agnostic meta-learning for fast adaptation of deep networks,'' in \emph{Proc. Int. Conf. Mach. Learn.}\hskip 1em plus 0.5em minus 0.4em\relax PMLR, 2017, pp. 1126--1135.

\bibitem{kingma2014adam}
D.~P. Kingma and J.~Ba, ``Adam: A method for stochastic optimization,'' \emph{arXiv preprint arXiv:1412.6980}, 2014.

\bibitem{BayesianOptimization}
\BIBentryALTinterwordspacing
F.~Nogueira, ``{Bayesian Optimization}: Open source constrained global optimization tool for {Python},'' 2014--. [Online]. Available: \url{https://github.com/bayesian-optimization/BayesianOptimization}
\BIBentrySTDinterwordspacing

\end{thebibliography}

\end{document}